%% file: pan-submission.tex
\documentclass[]{ceurart}

\usepackage[breakable,most]{tcolorbox}
\usepackage{annotation}
\usepackage{amsmath}
\usepackage{subcaption}

\usepackage{hyperref}
\usepackage{colortbl}
\definecolor{darkblue}{rgb}{0, 0, 0.5}
\definecolor{Lightgreen}{RGB}{144,238,144}
\definecolor{Gray}{gray}{0.9}

\usepackage{styles/modernbox}

\hypersetup{colorlinks=true, citecolor=darkblue, linkcolor=darkblue, urlcolor=darkblue}

\usepackage{booktabs}
\usepackage{multirow}
\usepackage{tabularx}
\usepackage{graphicx}

\begin{document}
\copyrightyear{2025}
\copyrightclause{Copyright for this paper by its authors.
  Use permitted under Creative Commons License Attribution 4.0
  International (CC BY 4.0).}

\conference{CLEF 2025 Working Notes, 9 -- 12 September 2025, Madrid, Spain}

\title{Overview of the Plagiarism Detection Task at PAN 2025}

\author[1,2]{Andr\'{e} Greiner-Petter}[%
orcid=0000-0002-5828-5497,
email=greinerpetter@gipplab.org,
url=https://gipplab.org/team/dr-andre-greiner-petter/,
]
\cormark[1]
\address[1]{Georg-August-Universit\"{a}t, G\"{o}ttingen, Germany}
\address[2]{National Institute of Informatics, Tokyo, Japan}

\author[3]{Maik Fr\"{o}be}[%
orcid=0000-0002-1003-981X,
email=pan@webis.de
]
\fnmark[1]
\address[3]{Friedrich-Schiller-Universit\"{a}t Jena, Jena, Germany}

\author[1]{Jan Philip Wahle}[%
orcid=0000-0002-2116-9767,
email=wahle@uni-goettingen.de
]
\fnmark[1]

\author[1]{Terry Ruas}[%
orcid=0000-0002-9440-780X,
email=ruas@gipplab.org
]
\fnmark[1]

\author[1]{Bela Gipp}[%
orcid=0000-0001-6522-3019,
email=gipp@gipplab.org
]

\author[2]{Akiko Aizawa}[%
orcid=0000-0001-6544-5076,
email=aizawa@nii.ac.jp
]

\author[4,5,6]{Martin Potthast}[%
orcid=0000-0003-2451-0665,
email=pan@webis.de,
url=pan.webis.de
]

\address[4]{University of Kassel, Kassel, Germany}
\address[5]{hessian.ai, Darmstadt, Germany}
\address[6]{ScaDS.AI, Leipzig, Germany}

\cortext[1]{Corresponding author.}
\fntext[1]{These authors contributed equally.}

\begin{abstract}
The generative plagiarism detection task at PAN 2025 aims at identifying automatically generated textual plagiarism in scientific articles and aligning them with their respective sources.
We created a novel large-scale dataset of automatically generated plagiarism using three large language models: Llama, DeepSeek-R1, and Mistral.
In this task overview paper, we outline the creation of this dataset, summarize and compare the results of all participants and four baselines, and evaluate the results on the last plagiarism detection task from PAN 2015 in order to interpret the robustness of the proposed approaches.
We found that the current iteration does not invite a large variety of approaches as naive semantic similarity approaches based on embedding vectors provide promising results of up to 0.8 recall and 0.5 precision.
In contrast, most of these approaches underperform significantly on the 2015 dataset, indicating a lack in generalizability.
\end{abstract}


\maketitle
\AddAnnotationRef{}

\input{clef25-pan-overview-task}

\begin{acknowledgments}
  This work has been funded by the Deutsche Forschungsgemeinschaft (DFG, German Research Foundation) -- 554559555, 564661959, 437179652; the Lower Saxony Ministry of Science and Culture, and the VW Foundation.
\end{acknowledgments}

\section*{Declaration on Generative AI}
  The author(s) have not employed any Generative AI tools.

\bibliography{clef25-pan-overview-lit,clef25-participant-notebooks-lit}

\input{annotation.tex}

\end{document}

%% file: clef25-pan-overview-task.tex
\section{Generative Plagiarism Detection}\label{generative-plagiarism-detection}
Plagiarism detection has a long-standing tradition at PAN, with the main tasks running from 2009~\cite{PotthastSEBR09} to 2015~\cite{StamatatosPPRS15}. 
Over time, the focus gradually shifted toward specialized intrinsic tasks, such as the still active authorship analysis challenges. 
However, the recent breakthrough of generative artificial intelligence (AI) has dramatically transformed the landscape of plagiarism detection. 
For the first time in history, large language models (LLMs) can serve as so-called automatic plagiarists~\cite{BarronCedenoPRS10}. 
At the same time, major scientific venues adjust their submission policies to allow (at least partially) AI-generated content~\cite{AclAiPolicy,AAAIAiPolicy,IcmlAiPolicy}.
The annual conference on AI (AAAI) recently announced to deploy an AI-assistend peer review assessment system for 2026\footnote{\url{https://aaai.org/aaai-launches-ai-powered-peer-review-assessment-system/}}.
This shift inspired us to revive a classic plagiarism detection task for 2025, this time centered on automatically generated plagiarism using LLMs.

For the 2025 edition, we adhered to the well-established foundations of the 2015 plagiarism detection task, particularly in evaluation methodology and dataset formatting~\cite{BarronCedenoPRS10}. Following the same formats will later allow us to evaluate new submissions on the older datasets to investigate the robustness of new approaches. 
Therefore, this format allows us to re-run the old baselines on this new dataset to judge the overall challenge of the new data versus the previous dataset.
The participants receive an annotated synthetic dataset of pairs of documents $(S, P)$, where $S$ is a source document and $P$ is the plagiarism document in which some paragraphs $p$ are replaced with paraphrased versions $s'$ of paragraphs $s$ in $S$ using an LLM without citation. 
This setup closely mirrors the 2015 PAN text alignment task\footnote{\url{http://www.uni-weimar.de/medien/webis/events/pan-15/pan15-web/plagiarism-detection.html}}.

The 2025 PAN task has received four submissions in total, outperforming all our baselines. 
Since all of these submissions (and our baselines) follow a similar approach of aligning text fragments based on their semantic similarity in terms of vector representations, we set up a fourth baseline using the Linq-Embed-Mistral model~\cite{linq}\footnote{in the following referred to as Linq}. 
Linq outperforms all submissions, indicating that specialized models for the text retrieval task might suit the task for plagiarism detection particularly well.
Note that this summary is an extended and in-depth version of the \textit{Overview of PAN 2025} paper~\cite{PAN:2025}.

\section{Dataset}
To the best of our knowledge, no large-scale dataset with automatically generated cases of textual reuse exists. 
Some studies suggest that LLMs can disguise plagiarism via paraphrasing the original source~\cite{WahleRMG21,WahleRKG22}. 
Additionally, LLMs have already been successfully used to replace human paraphrasing on scale~\cite{CeginSB23}. 
For this task revival, we aim to create a novel dataset with realistic cases of textual reuse disguised via automated paraphrasing. 
To make this dataset large enough to enable possible fine-tuning approaches, we automated the full dataset creation pipeline.

For this year's iteration, we focus on the text alignment task setup, i.e., we provide participants with pairs of source and plagiarized documents $(S, P)$ and the participants are asked to identify and align the LLM-generated, plagiarized paragraphs $s'$ in $P$ with their respective source paragraphs $s$ in $S$.

\subsection{Data Creation}
We use arXiv as the source corpus for our novel dataset.
Specifically, the ar5iv\footnote{\url{https://ar5iv.labs.arxiv.org/}} release from 2025 of arXiv. This dataset contains all arXiv documents in a structured HTML5 format, which allows us to avoid most parsing problems of identifying paragraph splits, author identifications, citations, and more.
We sample a subset of 100,000 documents with an even distribution across all arXiv categories (also known as archives), to ensure a wide variety of topics.
These 100,000 documents serve as candidates for $S$.
Afterwards, we use the SPECTER model~\cite{CohanFBDW20} to create document embeddings and identify the semantically most similar documents (in terms of cosine similarity) to each $S$.
This gives us 100,000 pairs of $(S, P)$.

For each document pair $(S, P)$, we first select a random number of paragraphs in $P$ that should be replaced with paragraphs from $S$.
Additionally, we add paragraphs $p$ that cite $S$ to the pool, as otherwise the document could contain genuine, referenced materials from $S$.
For each selected $p$, we than find the most semantically similar paragraphs $s$ based on three criteria. 
The alignment score is computed as a weighted aggregate: 50\% semantic similarity via SPECTER sentence embeddings, 40\% lexical similarity using TF-IDF vector similarity, and 10\% section title similarity using again SPECTER embeddings. 
The inclusion of similarity in the title of the section helps discourage the alignment of paragraphs from unrelated sections of the documents and preserve a more coherent document structure within $P$.
For each pair $(S, P)$, we select one of three LLMs: LLaMA-3~\cite{llama3modelcard} (3.3 70B Instruct), DeepSeek-R1~\cite{deepseekai2024deepseekv3technicalreport} (Distill-Qwen-32B) or Mistral~\cite{mistralmodelcard} (7B Instruct v0.3), and replace all selected $p$ in each aligned paragraph $(s, p)$ with LLM-paraphrased versions $s'$ derived from paragraphs $s$ in $S$.

\subsection{Categorization}

To support a more detailed analysis of system performances, we establish several categories of document pairs, which later allows us to slice the dataset and investigate performances (e.g., least recall) on specific subsets of the data. 
First, 5\% of the 100,000 pairs remain unchanged, i.e., both $S$ and $P$ are original arXiv documents without textual reuse. 
An additional 20\% of pairs do not contain any plagiarism, but some paragraphs in $P$ have been paraphrased by an LLM independently of $S$. 
These examples are useful for evaluating systems that aim to detect LLM-generated content rather than plagiarism specifically. 
We want to discourage such approaches, as the use of LLMs in modern research does not necessarily indicate academic misconduct or even plagiarism~\cite{Jarrah23}. 
Those document pairs are called \textbf{altered}.
The remaining 75\% of document pairs are constructed as plagiarism pairs as described above. 
In about half of these plagiarized documents, we also add 10\% of altered paragraphs so that plagiarized documents may also contain LLM-generated but otherwise genuine paragraphs.

\subsubsection{Severity.} We classify the severity of plagiarism in $P$ into three levels: low, medium, and high. 
These refer to the proportion of paragraphs in $P$ that are replaced with paraphrased versions from $S$. 
In 30\% of the document pairs, the severity is \textit{low}, with 20\% to 40\% of paragraphs replaced. 
In 40\% of the pairs, severity is \textit{medium}, with 40\% to 60\% replaced. 
The remaining 30\% has \textit{high} severity, where 70\% to 100\% of paragraphs in $P$ are substituted.

\subsubsection{Paraphrasing Prompts.}
For paraphrasing, we use three prompt types: simple, default, and complex.
While severity is defined on a document pair level, each pair of paragraphs within one document pair can use different types of prompts. 
For each pair, we follow a distribution of 60\% simple prompts, 30\% default prompts, and 10\% complex prompts.
The \textit{simple prompt} instructs the LLM to paraphrase a given paragraph without additional constraints.
\begin{llmpromptbox}{Simple Paraphrasing Prompt}
\texttt{Paraphrase the given paragraph for a professional audience.}
\end{llmpromptbox}

We found that, especially technical texts, like the ones we often find in scientific articles from arXiv, do not produce sufficient paraphrasing. 
This is especially prominent to see if the texts contain mathematical formulae.
To encourage the LLMs to generate more sophisticated paraphrasing, we use different \textit{default prompt} that elevates the use of a complete reformulation rather than slight adjustments.
\begin{llmpromptbox}{Default Paraphrasing Prompt}
\texttt{Reformulate the given paragraph in a sophisticated manner while preserving its meaning. Modify sentence structure, reword phrases, and incorporate elements of general knowledge to ensure coherence. The less token overlap, the better.}
\end{llmpromptbox}

As the synthetic data faces the issue of replacing paragraphs from an existing, genuine document, one could potentially identify incoherent logical steps from one paragraph to the other in order to identify replaced paragraphs. 
In order to make this a more realistic setup, we define a third type of prompt that tries to take the previous paragraph into account as a context for the LLM to generate slightly more appropriate paraphrasing. 
\begin{llmpromptbox}{Complex Paraphrasing Prompt Structure with Context}
\texttt{Completely rephrase the given paragraph in your own words. Feel free to incorporate elements from general knowledge to ensure coherence, flow, and better understanding.} \\ \\
\texttt{\{context\_before\}}
\end{llmpromptbox}

All prompts include additional instructions to output only the paraphrased content, avoiding any explanatory text. 
Special tokens are used to suppress verbose output, tailored to each LLM. 
For DeepSeek-R1, a custom \texttt{<thinking>\dots</thinking>} block was used to suppress the model’s internal reasoning steps, which would otherwise significantly slow down the generation.
It is worth noting that Mistral performed poorly in following prompt instructions. 
It often produces explanatory content, hallucinated facts, or gets stuck in output loops, an issue reminiscent of neural network architectures before the attention mechanism era~\cite{FuLSS21}. 
We presume the 7B parameter model variant is simply too small to perform paraphrasing of highly technical texts.
In total, the final dataset consists of 78,038 document pairs, divided into training, validation, and test subsets. 
The training and validation sets are provided to participants, while the test set is kept private for the evaluation phase. 
The data splits and sizes are given in Table~\ref{table-plag-data}.

\begin{table}[t!]
\small
\setlength{\tabcolsep}{4pt}
\centering
\caption{Plagiarism alignment dataset and LLM splits. The bottom percentages refer to the total amount of samples in the final dataset.}
\label{table-plag-data}
\resizebox{\textwidth}{!}{
\begin{tabular}{l|rrrrrr|r|r|r}
\toprule
 & \multicolumn{2}{c}{\textbf{Llama-3}} & \multicolumn{2}{c}{\textbf{DeepSeek-R1}} & \multicolumn{2}{c|}{\textbf{Mistral}} & \textbf{Alt.} & \textbf{Orig.} & \textbf{Total} \\
\midrule
Train      & 18,423 & 79.80\% & 18,452 & 79.46\% & 6,265 & 79.65\% & 15,101 & 3,918 & 62,159 \\
Validation &  2,353 & 10.19\% &  2,383 & 10.26\% &   802 & 10.20\% &  1,919 &   518 &  7,975 \\
Test       &  2,310 & 10.01\% &  2,386 & 10.28\% &   799 & 10.16\% &  1,919 &   490 &  7,904 \\
\midrule
\textbf{Total} & 23,086 & 42.62\% & 23,221 & 42.86\% & 7,866 & 14.52\% & 18,939 & 4,926 & 78,038 \\
\bottomrule
\end{tabular}
}
\end{table}

\section{Evaluation}
All systems are submitted and evaluated on the TIRA platform~\cite{froebe:2023b}. 
The participants are tasked with identifying all the paragraphs $s'$ in $P$ and aligning each with the corresponding paragraph $s$ in $S$.
The training and validation sets contain all alignments $(s, s')$ for each pair of documents $(S, P)$, together with the full text of both documents. 
The evaluation is carried out using the original scripts from the 2015 PAN plagiarism detection task. 
We used granularity as well as the micro-averaged and macro-averaged variants of \texttt{plagdet}, recall, and precision for comparability purposes with past plagiarism detection tasks~\cite{PotthastSBR10}.
%
All of these metrics take into account the exact character spans of the source and plagiarism and calculate the overlap regions in comparison to the truth values.
While the micro-averaged variants take the length of plagiarism spans into account, the macro-averaged variants are length independent.
The micro-averaged variants made especially sense for the old task setups at PAN, as earlier iterations infused plagiarism on sentence and sometimes even subsentence levels.
As our dataset is constructed based on paragraph borders, the micro-variants are less indicative for our evaluations.
For the sake of completeness, we evaluated all algorithms on both variants.

The granularity metric counts how often a truth case is detected on average. 
This metric is useful as we want to avoid a single case of plagiarism being detected multiple times.
The domain of the granularity metric is $[ 1, |D| ]$ where $|D|$ is the number of detections for a single document pair.
A perfect score of $1$ means that every truth case of plagiarism is detected at most once by the given algorithm.
As a reminder, \texttt{plagdet} is defined via the $F_1$ score and with respect to the granularity:
\begin{equation}
    \texttt{plagdet} (P, D) = \frac{F_1 (P, D)}{\log_2 \left( 1+\text{gran}(P, D) \right)},
\end{equation}
where $P$ indicates the actual case of plagiarism in the truth data and $D$ the detected cases in $(S, P)$. 

\subsection{Baselines}
We implement three new baselines that use semantic similarity with large language models and the baseline from the 2012~edition of PAN~\cite{potthast:2012} that uses lexical similarity. For the three large language model baselines, we split $S$ and $P$ into their paragraphs. For each paragraph in $p$ we take the semantically closest paragraph in $S$ in terms of cosine similarity based on \textit{Linq}~\cite{linq}, \textit{Qwen2 7B instruct}\footnote{In the following referred to as Qwen2.}~\cite{qwen}, and \textit{Llama-3.3 70B Instruct}\footnote{In the following referred to as Llama}~\cite{llama3modelcard}. For each model, we define a cut-off threshold that classifies the closest pairs as plagiarism. 
Pairs below that threshold are then discarded. 
The threshold is determined by calculating the ideal cut-offs on the training split of the data. To compare this class of semantic plagiarism detectors to previous lexical approaches, we also include the baseline from the 2012~edition of the plagiarism detection task at PAN. The 2012 baseline tokenizes the text while normalizing white spaces and punctuation and then detects sequences of overlapping n-grams between $S$ and $P$ as plagiarism cases.

\subsection{Team Submissions}
Four teams participated in the task by submitting software. 

\subsubsection{Team chi-zi-zhi-xin-dui.} Su et al.~\cite{su:2025} split the document of each pair into sentences and aligned the sentences of $S$ and $P$ according to the SBERT, MPNet, TF-IDF, or BERT score, whichever passed a pre-defined threshold, which was also determined based on the training data. 
After the alignment, they performed a merging logic to combine subsequences of detected sentences into single blocks.

\subsubsection{Team foshan-university.} Tang et al.~\cite{tang:2025} also pre-processed documents by splitting them into sentence chunks and aligned all sentences from $P$ with sentences from $S$ based on E5 embeddings (\texttt{intfloat/e5-base-v2}).
Again, the threshold was determined with the training data. 
They also performed a span aggregation if two spans have been categorized as plagiarism within a distance of 30 characters.

\subsubsection{Team jrluo.} Jieren et al.~\cite{jieren:2025} also split the documents into sentences and first aligned pairs by using TF-IDF vector similarities. 
For each pair, he calculated the word-based Jaccard similarity and discarded all pairs below a given threshold. 
All remaining sentence pairs were classified as plagiarism or genuine by a BERT classifier fine-tuned on the training data.

\subsubsection{Team yukino.} Mo et al.\cite{mo:2025} also splits the data into chunks of sentences. Each sentence gets a vector representation as the averaged vector representation of each token based on Glove (6B model with 300 dimensions). 
Afterwards, all sentences are aligned according to their cosine similarities. Like all other teams, Mo et al.~also employed a merging strategy for positive detections based on position proximity, semantic coherence (based on cosine similarity), and a minimum length constraint.

\subsection{Discussion and Results}

\begin{table}[t]
\setlength{\tabcolsep}{6pt}
\centering
\caption{Overall results of baselines and submissions. $r$ and $p$ refer to recall and precision, respectively. Sore is the mean average of all individual scores.}
\label{table-plag-ranking}
\resizebox{\textwidth}{!}{
\begin{tabular}{l|ccc|ccc|r|c}

\toprule
 & \multicolumn{3}{c|}{Micro} & \multicolumn{3}{c|}{Macro} &  &  \\

\textbf{submission} & $plagdet$ & $r$ & $p$ & $plagdet$ & $r$ & $p$ & gran. & score \\
\midrule
qwen2 & $0.39$ & $0.57$ & $0.32$ & $0.26$ & $0.54$ & $0.18$ & $1.06$ & $0.38$ \\
linq & $\bf 0.61$ & $\bf 0.82$ & $0.58$ & $\bf 0.53$ & $\bf 0.83$ & $0.45$ & $1.15$ & $\bf 0.64$ \\
llama & $0.28$ & $0.36$ & $0.23$ & $0.21$ & $0.40$ & $0.14$ & $1.01$ & $0.27$ \\
pan12 & $0.08$ & $0.08$ & $0.59$ & $0.06$ & $0.05$ & $\bf 0.54$ & $2.20$ & $0.23$ \\
\midrule
foshan-university & $0.42$ & $0.57$ & $0.33$ & $0.31$ & $0.56$ & $0.21$ & $\bf 1.00$ & $0.40$ \\
jrluo & $0.16$ & $0.12$ & $0.53$ & $0.14$ & $0.10$ & $0.53$ & $1.39$ & $0.26$ \\
chi-zi-zhi-xin-dui & $0.42$ & $0.38$ & $\bf 0.67$ & $0.34$ & $0.32$ & $0.51$ & $1.21$ & $0.44$ \\
yukino & $0.49$ & $0.50$ & $0.48$ & $0.45$ & $0.46$ & $0.45$ & $\bf 1.00$ & $0.47$ \\

\bottomrule
\end{tabular}
}
\end{table}

Table~\ref{table-plag-ranking} shows the evaluation results for all submissions and baselines on our new dataset. 
The final score is the average of all sub-scores and is reported as the final score in the lab overview paper~\cite{PAN:2025}. 
While Linq seems to outperform most other approaches, the best performers vary in terms of precision and granularity. 
This is especially surprising as the baselines Linq, Qwen2, and Llama have been deployed for paragraph splitting rather than sentence splitting with subsequent merging techniques. 
We would assume these baselines have a slight advantage, especially on the granularity score. 
It should also be noted that Linq was deployed afterwards to investigate the performance of a special model that aimed towards text retrieval tasks. 
Otherwise, most submissions outperform the baselines with the exception of team jrluo.
Team jrluo has a relatively low recall compared to high precision scores. 
We suspect this is related to an agressive filtering of the initial TF-IDF similarity calculations.

\begin{table}[t]
\setlength{\tabcolsep}{6pt}
\centering
\caption{Overall results of baselines and submissions on the old PAN12 dataset. $r$ and $p$ refer to recall and precision, respectively. Sore is the mean average of all individual scores.}
\label{table-plag-ranking-pan12}
\resizebox{\textwidth}{!}{
\begin{tabular}{l|ccc|ccc|r|c}

\toprule
 & \multicolumn{3}{c|}{Micro} & \multicolumn{3}{c|}{Macro} &  &  \\

\textbf{submission} & $plagdet$ & $r$ & $p$ & $plagdet$ & $r$ & $p$ & gran. & score \\
\midrule
qwen2 & $0.03$ & $\bf 0.49$ & $0.04$ & $0.01$ & $0.38$ & $0.02$ & $7.49$ & $0.16$ \\
linq & $0.17$ & $0.45$ & $0.71$ & $0.14$ & $\bf 0.42$ & $0.44$ & $7.99$ & $0.39$ \\
llama & $0.02$ & $0.34$ & $0.05$ & $0.01$ & $0.19$ & $0.02$ & $10.94$ & $0.10$ \\
pan12 & $\bf 0.29$ & $0.35$ & $\bf 0.99$ & $\bf 0.22$ & $0.24$ & $0.93$ & $2.38$ & $\bf 0.50$ \\
\midrule
foshan-university & $0.03$ & $0.03$ & $0.03$ & $0.03$ & $0.04$ & $0.02$ & $\bf 1.04$ & $0.03$ \\
jrluo & $0.08$ & $0.10$ & $0.98$ & $0.08$ & $0.10$ & $\bf 0.97$ & $3.69$ & $0.38$ \\
chi-zi-zhi-xin-dui & $0.01$ & $0.01$ & $0.22$ & $0.02$ & $0.02$ & $0.08$ & $2.38$ & $0.06$ \\

\bottomrule
\end{tabular}
}
\end{table}

Table~\ref{table-plag-ranking-pan12} shows the same results on the old PAN12 dataset. 
Unfortunately, team yukino could not be evaluated as we ran into issues when applying the old datasets. 
All submissions (except the original PAN12 baseline) face a significant drop in performance.
This is not as surprising for the baselines, as the paragraph splitting simply should not have been applied to the old dataset. 
This is also evident when looking at the high granularity scores.
The team submissions perform significantly better in terms of granularity.
An outlier is again team jrluo with very high precision values.
It seems the two-stage filtering approach is particularly useful on the older dataset. 

\begin{figure}[t]
    \centering
    \begin{subfigure}{0.49\linewidth}  
        \centering
        \includegraphics[width=\textwidth]{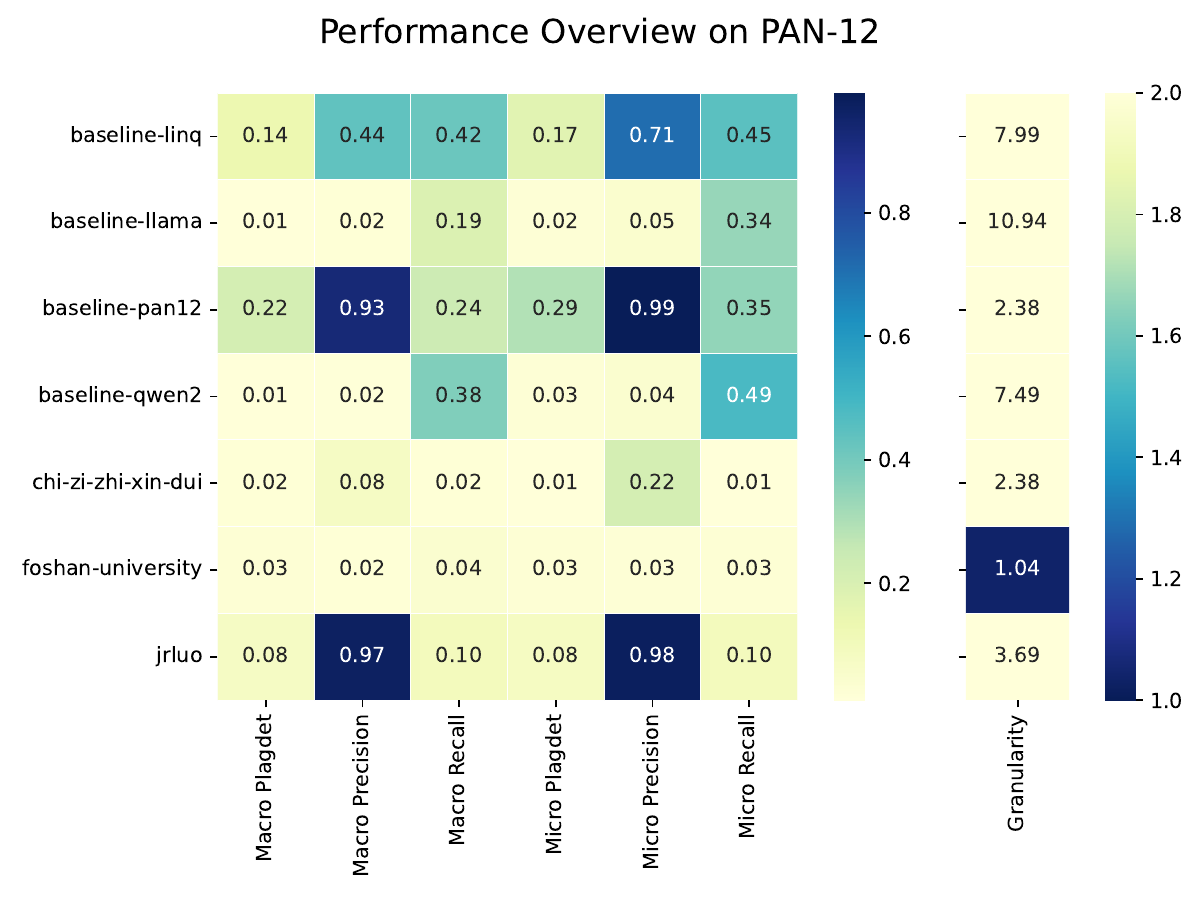}
        \label{fig:heatmap-pan12}
    \end{subfigure}
    \hfill  
    \begin{subfigure}{0.49\linewidth}  
        \centering
        \includegraphics[width=\textwidth]{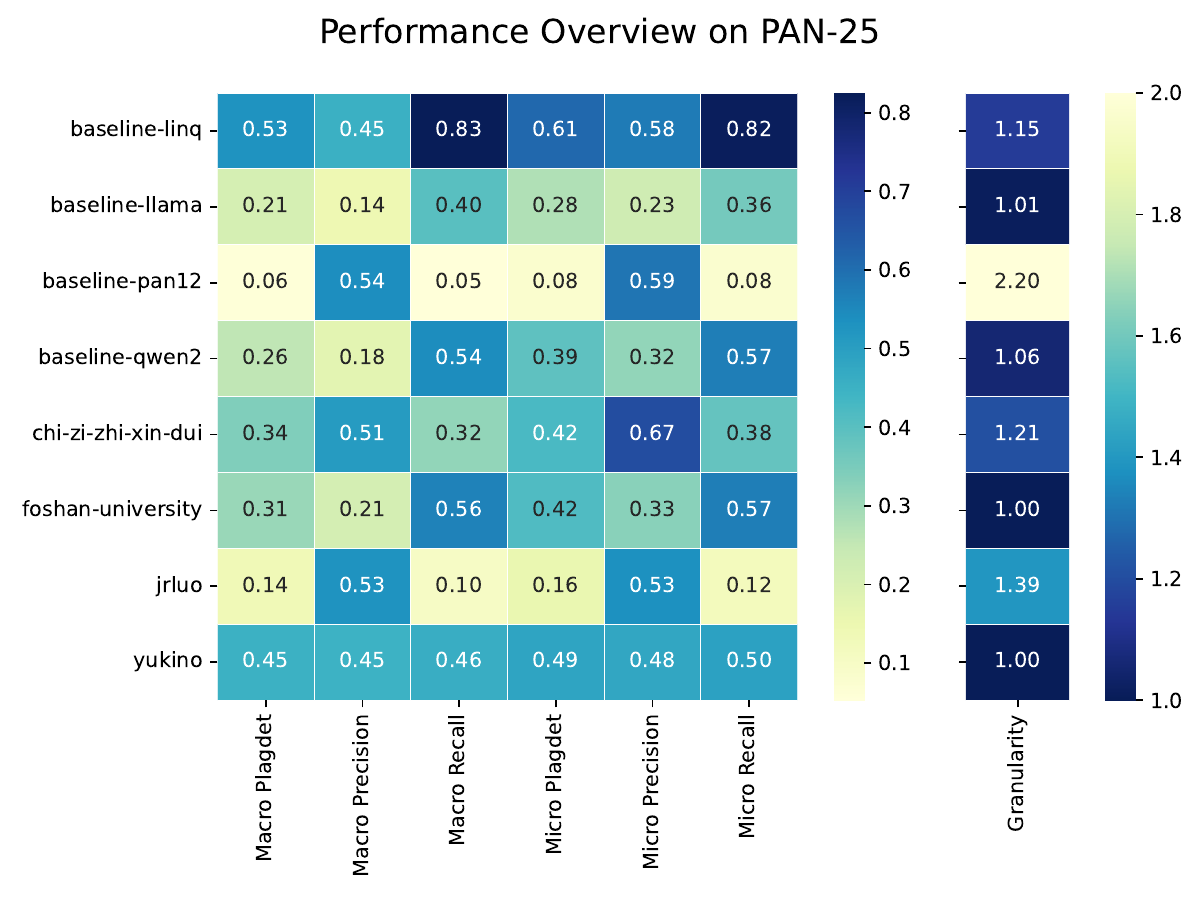}
        \label{fig:heatmap-pan25}
    \end{subfigure}

    \caption{Comparison heatmap of performances on the PAN-12 and PAN-25 datasets.}
    \label{fig:heatmap-comparison}
\end{figure}

Figure~\ref{fig:heatmap-comparison} shows the results as a heatmap layout. 
We can see that team yukino performs overall similarly to Linq but loses significant on recall. 
It is also noteworthy that the new dataset is significantly easier in terms of granularity, as entire paragraphs have been plagiarized. It is therefore relatively rare that multiple detections detect the same plagiarized paragraph.

\subsubsection{Data Subsets.}
\begin{figure}[t]
    \centering
    \begin{subfigure}{0.49\linewidth}  
        \centering
        \includegraphics[width=\textwidth]{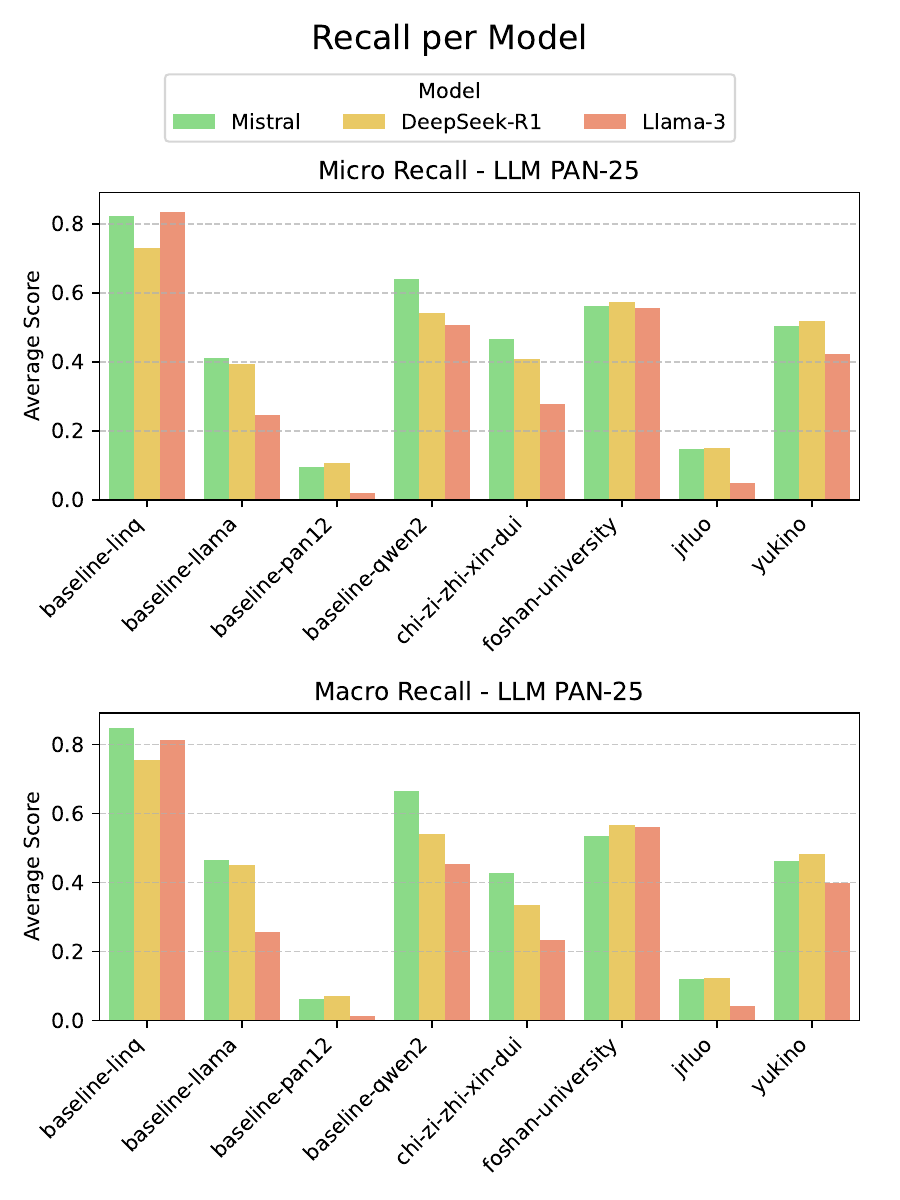}
        \label{fig:modeltypes}
    \end{subfigure}
    \hfill 
    \begin{subfigure}{0.49\linewidth}  
        \centering
        \includegraphics[width=\textwidth]{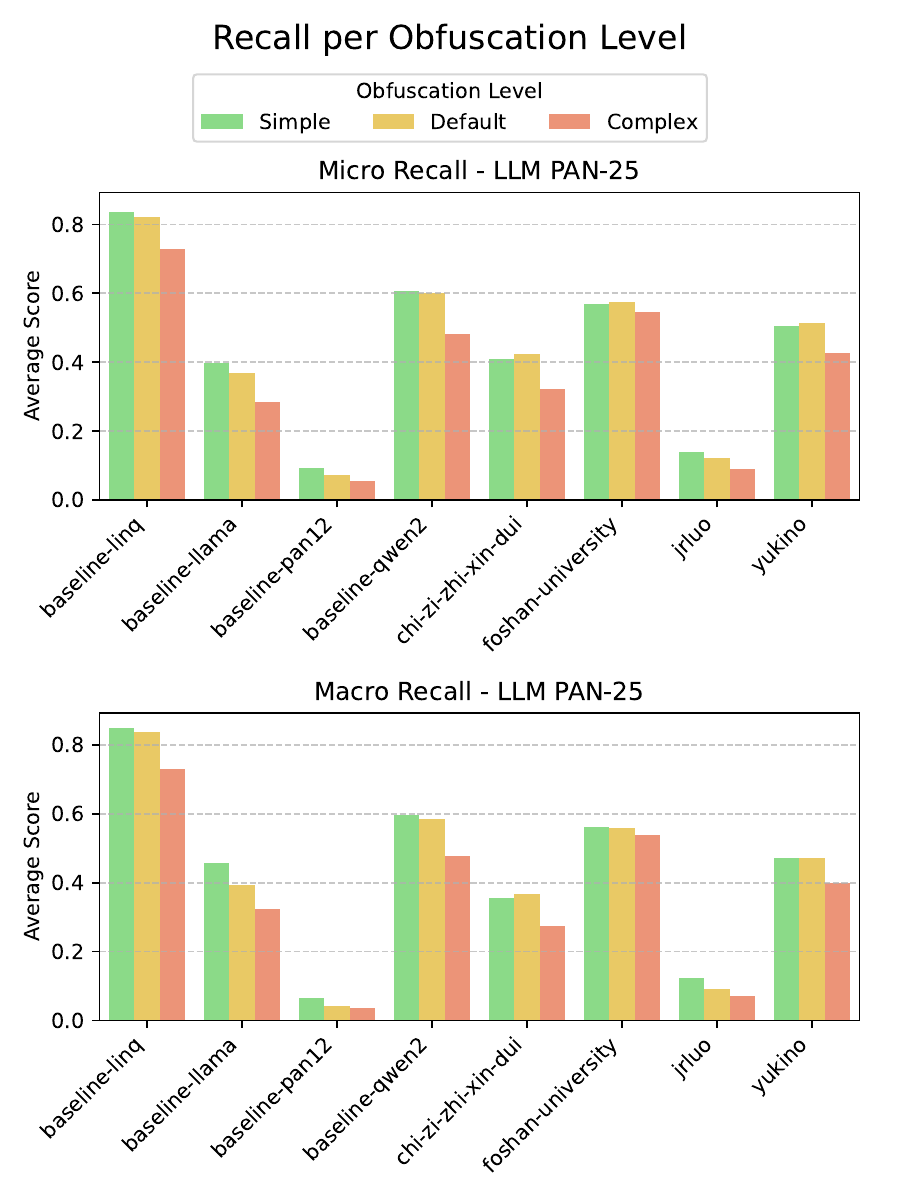}
        \label{fig:prompts}
    \end{subfigure}

    \caption{Comparison of recall values per used models and used prompts.}
    \label{fig:prompt-model}
\end{figure}

To investigate the performance on specific subsets of the data, we calculate the recall values on slices of the data.
We only calculate the recall metrics of all approaches on the new data, as precision, plagdet, and granularity would require us to rerun all submissions on a pre-filtered dataset. 
However, the recall performances are sufficient to identify trends across all submissions.
Figure~\ref{fig:prompt-model} shows the recall values of all algorithms on different models (i.e., which model is used to generate the paraphrasing) and the obfuscation level (i.e., which prompt has been used to generate the paraphrasing). 

We can see that overall, Mistral is easier to detect by almost all approaches. 
A possible explanation is the fact that Mistral, with 7B parameters, is also the smallest of the used models and as such did not provide as high-quality paraphrasing as the other models did.
Counterintuitively, the Llama baseline performs significantly worse on detecting paraphrasing generated with Llama.
This might be surprising as classical LLM-detection methods typically perform best when the same model was used for the detection as for the generation of the texts~\cite{Mitchell0KMF23,abs-2502-12743,TangCH24}.
However, the results should not be confused with classical LLM-detection approaches that are often based on logit-value comparisons.
The approaches here are marely based on cosine similarities of content embeddings rather than logit values between tokens.

Another trend is also visible in the obfuscation level overview.
The recall values per obfuscation level confirm a clear difference between prompt types.
Almost all approaches find more plagiarism generated with simple prompts.
Likewise, all approaches have the lowest success detection rate with complex prompts.
While some approaches, such as by {\it chi-zi-zhi-xin-dui}, are more susceptible to model changes, some approaches are relatively stable regardless of prompt or model type, such as {\it foshan-university}.

\begin{figure}[t]
    \centering
    \includegraphics[width=0.7\linewidth]{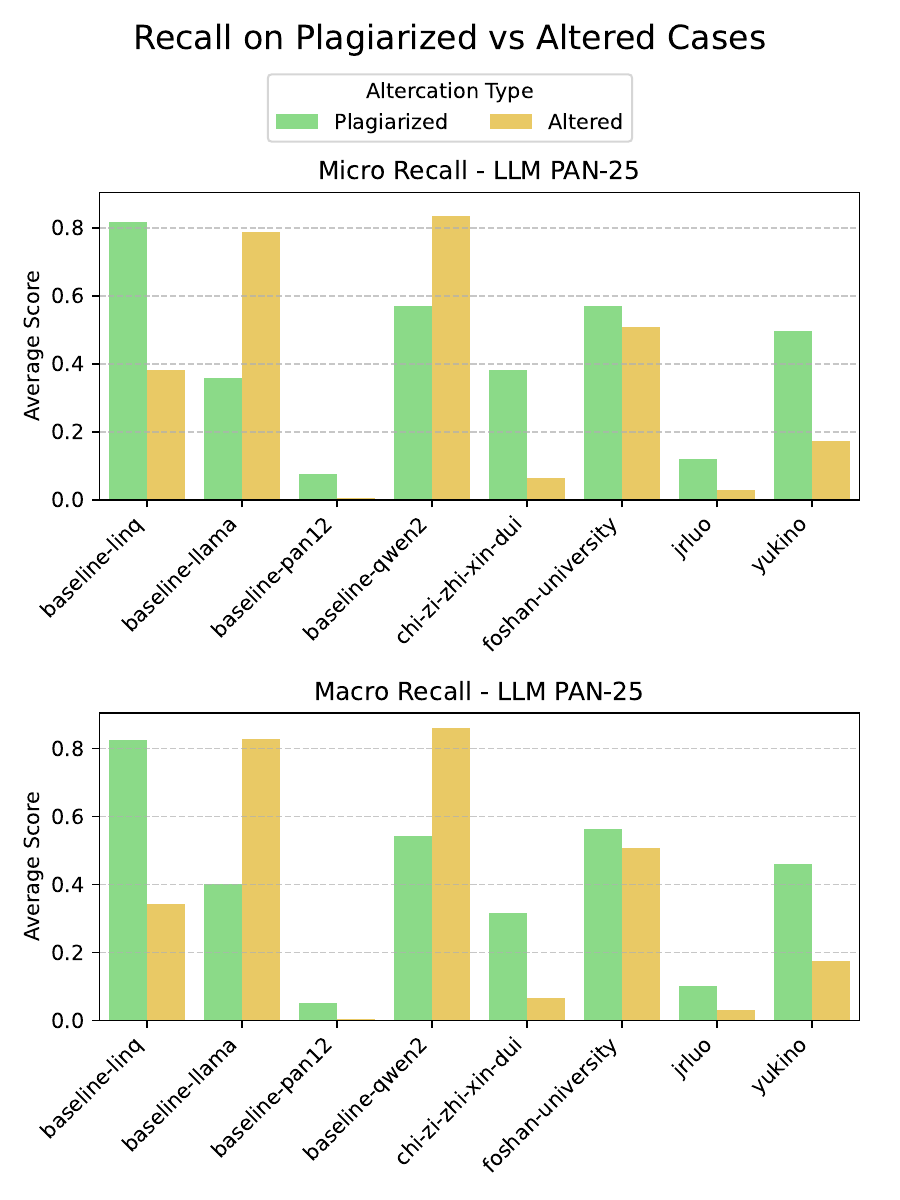}
    \caption{Recall on plagiarism cases versus altered genuine cases.}
    \label{fig:altered}
\end{figure}

Lastly, Figure~\ref{fig:altered} shows the recall performances on the actual plagiarism cases compared to all altered cases. 
Detecting an altered case is considered a false-positive.
We want approaches that minimize these false classifications, as they could be interpreted as potentially harmful false accusations when handling plagiarism detections.
Surprisingly, We can identify a clear difference between participant's submissions and two of our baselines even though the underlying approaches are not particularly diverse.
We can see that all submissions by participants show a significantly lower recall on altered cases, sometimes up to 20\% lower.
The baselines of Llama and Qwen2 are particularly noteworthy as opposing approaches. as the recall on altered cases is significantly higher (in the case of Llama, even twice as high) than on actual plagiarized cases.
That means, an identified case of plagiarism with these approaches is significantly more likely to be a wrong accusation than an actual case of plagiarism.
We assume this discrepancy comes from the construction of the dataset, as all pairs $(S, P)$ have been constructed to be semantically close.
We can therefore assume a relatively high, general similarity across all paragraphs between $S$ and $P$ even without infused plagiarism.
It seems Llama and Qwen2 have particular issues with differentiating these nuances in semantic similarities based on these embeddings.

In summary, the results mostly underperform our expactations. 
All submitted approaches and baselines follow a simple detection approach based on cosine similarities of content embeddings and achieve mostly values below $0.6$ in plagdet.
In comparison, on the 2014 edition of the text alignment task\footnote{\url{https://pan.webis.de/clef14/pan14-web/text-alignment.html}} the majority of submissions achieved plagdet scores above $0.8$.
Unfortunately, it is unclear if this can be attributed to a more difficult task setup or the simplicity of detection approaches.
The comparison to the the PAN12 dataset indicates that all approaches are not robust against changes in the data.
However, this also includes the previous PAN12 baseline as it outperforms other methodologies on the PAN12 task but significantly underperforms on the new dataset.

\section{Future Work}
The revival of the plagiarism detection task can be summarized as successful. 
However, there are a few crucial improvements that can be made to make this task more realistic.
The main point of criticism is the actual generation of plagiarism in the new dataset.
The current pipeline starts with two genuine documents and infuses synthetic plagiarism by replacing a subset of paragraphs with a paraphrased version of another article. 
Typically, the textual content of scientific articles is not that interchangeable.
Likewise, real-world plagiarism typically does not start with an existing publication and adds paragraphs from other works to it.
In order to overcome this issue, in future iterations, we will start from multiple genuine documents (or a single document) and generate a new article by paraphrasing the content of each source rather than replacing paragraphs within an existing document excerpt.
This should also promote a larger variety of detection approaches, as all submissions have been following very similar approaches.
The new pipeline will also allow us to revive the important retrieval aspect of plagiarism detection tasks, in which participants start from a suspicious document without knowing if it is genuine or what the sources are.
Another shortcoming is the relatively narrow domain of arXiv. 
As we have seen with the evaluations on the PAN12 dataset, all approaches, including the PAN12 baseline, are not very robust and perform vastly different on different datasets.
This means newer iterations of this task must incorporate a larger variety of types and possibly domain of plagiarism.
In the future, we will incorporate especially the medical domain to bring more variety to the dataset.

Another challenge is the rapid development of LLMs and plagiarism in itself.
Recently, Zochi, a scientific LLM has generated a publication that passed the scrutiny of peer reviews at a reputable international conference\footnote{\url{https://www.intology.ai/blog/zochi-acl}}.
This shows that LLMs are capable of generating genuine, new scientific texts without plagiarizing existing work.
Nonetheless, plagiarizing existing work is now easier than ever for perpetrators.
Future iterations of this task must therefore focus more on proper citations and the actual case of idealogical reuse or copying of reasoning-chains to stay relevant.
Proper citation of $S$ in $P$ was only touched on the surface in the creation of this iteration's dataset and not separately evaluated.
Lastly, this development also deemphasizes the alignment task because, moving forward, there will be fewer straightforward cases of matching sources to plagiarism. 
Instead, indicators such as structural, ideological, or reasoning chain similarities will have to be utilized to detect plagiarism.
We will therefore reframe future iterations of this task to ensure that the dataset and plagiarism detection approaches stay relevant regardless of the development of LLMs.

%% file: annotation.tex
\onecolumn
\hypertarget{annotation}{}
\pagestyle{empty}
\lstset{
  basicstyle=\footnotesize\ttfamily,
  breaklines=true,
  breakatwhitespace=false,
  columns=flexible,
  numbers=none
}

\definecolor{Primary}{RGB}{59, 130, 246}    
\definecolor{PrimaryDark}{RGB}{30, 64, 175} 
\definecolor{LightBg}{RGB}{239, 246, 255}   
\definecolor{TextDark}{RGB}{31, 41, 55}     
\definecolor{TextMuted}{RGB}{107, 114, 128} 

\begin{tikzpicture}[remember picture, overlay]
  \fill[Primary] ([xshift=0cm,yshift=0cm]current page.north west) rectangle ([xshift=\paperwidth,yshift=-0.4cm]current page.north west);
\end{tikzpicture}

\vspace{0.8cm}
\begin{center}
  {\fontsize{22}{26}\selectfont\sffamily\bfseries \textcolor{PrimaryDark}{CiteAssist}}\\[0.2em]
  {\Large\sffamily\scshape \textcolor{TextMuted}{Citation Sheet}}\\[0.8em]
  {\small\sffamily Generated with \href{https://citeassist.uni-goettingen.de/}{\textcolor{Primary}{\texttt{citeassist.uni-goettingen.de}}}
  }\end{center}

\begin{center}
\vspace{1em}
\begin{tikzpicture}
\draw[Primary, line width=0.6pt] (0,0) -- (\textwidth,0);
\end{tikzpicture}
\vspace{1.2em}
\end{center}

\begin{tcolorbox}[enhanced,
                 frame hidden,
                 boxrule=0pt,
                 borderline west={2pt}{0pt}{Primary},
                 colback=LightBg,
                 sharp corners,
                 breakable,
                 fonttitle=\sffamily\bfseries\large,
                 coltitle=Primary,
                 title=BibTeX Entry,
                 attach title to upper={\vspace{0.2em}\par},
                 left=12pt]
\lstset{
    inputencoding = utf8,  
    extendedchars = true,  
    literate      =        
      {á}{{\'a}}1  {é}{{\'e}}1  {í}{{\'i}}1 {ó}{{\'o}}1  {ú}{{\'u}}1
      {Á}{{\'A}}1  {É}{{\'E}}1  {Í}{{\'I}}1 {Ó}{{\'O}}1  {Ú}{{\'U}}1
      {à}{{\`a}}1  {è}{{\`e}}1  {ì}{{\`i}}1 {ò}{{\`o}}1  {ù}{{\`u}}1
      {À}{{\`A}}1  {È}{{\`E}}1  {Ì}{{\`I}}1 {Ò}{{\`O}}1  {Ù}{{\`U}}1
      {ä}{{\"a}}1  {ë}{{\"e}}1  {ï}{{\"i}}1 {ö}{{\"o}}1  {ü}{{\"u}}1
      {Ä}{{\"A}}1  {Ë}{{\"E}}1  {Ï}{{\"I}}1 {Ö}{{\"O}}1  {Ü}{{\"U}}1
      {â}{{\^a}}1  {ê}{{\^e}}1  {î}{{\^i}}1 {ô}{{\^o}}1  {û}{{\^u}}1
      {Â}{{\^A}}1  {Ê}{{\^E}}1  {Î}{{\^I}}1 {Ô}{{\^O}}1  {Û}{{\^U}}1
      {œ}{{\oe}}1  {Œ}{{\OE}}1  {æ}{{\ae}}1 {Æ}{{\AE}}1  {ß}{{\ss}}1
      {ẞ}{{\SS}}1  {ç}{{\c{c}}}1 {Ç}{{\c{C}}}1 {ø}{{\o}}1  {Ø}{{\O}}1
      {å}{{\aa}}1  {Å}{{\AA}}1  {ã}{{\~a}}1  {õ}{{\~o}}1 {Ã}{{\~A}}1
      {Õ}{{\~O}}1  {ñ}{{\~n}}1  {Ñ}{{\~N}}1  {¿}{{?\`}}1  {¡}{{!\`}}1
      {„}{\quotedblbase}1 {“}{\textquotedblleft}1 {–}{$-$}1
      {°}{{\textdegree}}1 {º}{{\textordmasculine}}1 {ª}{{\textordfeminine}}1
      {£}{{\pounds}}1  {©}{{\copyright}}1  {®}{{\textregistered}}1
      {«}{{\guillemotleft}}1  {»}{{\guillemotright}}1  {Ð}{{\DH}}1  {ð}{{\dh}}1
      {Ý}{{\'Y}}1    {ý}{{\'y}}1    {Þ}{{\TH}}1    {þ}{{\th}}1    {Ă}{{\u{A}}}1
      {ă}{{\u{a}}}1  {Ą}{{\k{A}}}1  {ą}{{\k{a}}}1  {Ć}{{\'C}}1    {ć}{{\'c}}1
      {Č}{{\v{C}}}1  {č}{{\v{c}}}1  {Ď}{{\v{D}}}1  {ď}{{\v{d}}}1  {Đ}{{\DJ}}1
      {đ}{{\dj}}1    {Ė}{{\.{E}}}1  {ė}{{\.{e}}}1  {Ę}{{\k{E}}}1  {ę}{{\k{e}}}1
      {Ě}{{\v{E}}}1  {ě}{{\v{e}}}1  {Ğ}{{\u{G}}}1  {ğ}{{\u{g}}}1  {Ĩ}{{\~I}}1
      {ĩ}{{\~\i}}1   {Į}{{\k{I}}}1  {į}{{\k{i}}}1  {İ}{{\.{I}}}1  {ı}{{\i}}1
      {Ĺ}{{\'L}}1    {ĺ}{{\'l}}1    {Ľ}{{\v{L}}}1  {ľ}{{\v{l}}}1  {Ł}{{\L{}}}1
      {ł}{{\l{}}}1   {Ń}{{\'N}}1    {ń}{{\'n}}1    {Ň}{{\v{N}}}1  {ň}{{\v{n}}}1
      {Ő}{{\H{O}}}1  {ő}{{\H{o}}}1  {Ŕ}{{\'{R}}}1  {ŕ}{{\'{r}}}1  {Ř}{{\v{R}}}1
      {ř}{{\v{r}}}1  {Ś}{{\'S}}1    {ś}{{\'s}}1    {Ş}{{\c{S}}}1  {ş}{{\c{s}}}1
      {Š}{{\v{S}}}1  {š}{{\v{s}}}1  {Ť}{{\v{T}}}1  {ť}{{\v{t}}}1  {Ũ}{{\~U}}1
      {ũ}{{\~u}}1    {Ū}{{\={U}}}1  {ū}{{\={u}}}1  {Ů}{{\r{U}}}1  {ů}{{\r{u}}}1
      {Ű}{{\H{U}}}1  {ű}{{\H{u}}}1  {Ų}{{\k{U}}}1  {ų}{{\k{u}}}1  {Ź}{{\'Z}}1
      {ź}{{\'z}}1    {Ż}{{\.Z}}1    {ż}{{\.z}}1    {Ž}{{\v{Z}}}1  {ž}{{\v{z}}}1
  }
\begin{lstlisting}
@inproceedings{greinerpetter25,
  author={Greiner-Petter, Andre and Froebe, Maik and Wahle, Jan Philip and Ruas, Terry and Gipp, Bela and Aizawa, Akiko and Potthast, Martin},
  title={Overview of the Plagiarism Detection Task at PAN 2025},
  booktitle={16th International Conference of the CLEF Association (CLEF)},
  publisher={CEUR-WS.org},
  series={CEUR Workshop Proceedings},
  topic={pd},
  volume={4038},
  year={2025},
  month={12}
}
\end{lstlisting}
\end{tcolorbox}

\vfill
\begin{tikzpicture}
\draw[Primary!40, line width=0.4pt] (0,0) -- (\textwidth,0);
\end{tikzpicture}
\begin{center}
\small\sffamily\textcolor{TextMuted}{Generated \today}
\end{center}